\documentclass[copyright,creativecommons]{eptcs} 
\providecommand{\event}{ICLP 2025} 

\usepackage{iftex}

\usepackage{listings}
\usepackage[inline]{enumitem} 
\usepackage{url}

\usepackage{amssymb}

\newcommand{\func}[1]{\ensuremath{\mathsf{#1}}}

\newcommand{\variablesymbols}{\ensuremath{\mathcal{V}}}
\newcommand{\constantsymbols}{\ensuremath{\mathcal{C}}}
\newcommand{\functionsymbols}{\ensuremath{\mathcal{F}}}
\newcommand{\predicatesymbols}{\ensuremath{\mathcal{P}}}
\newcommand{\folanguage}{\ensuremath{\langle \variablesymbols, \constantsymbols, \functionsymbols, \predicatesymbols \rangle}}
\newcommand{\prog}{\ensuremath{P}}
\newcommand{\AS}{\ensuremath{ASet}}
\newcommand{\universe}{\ensuremath{U}}
\newcommand{\base}{\ensuremath{B}}
\newcommand{\interpretation}{\ensuremath{I}}

\newcommand{\nafsymbol}{\ensuremath{not}}
\newcommand{\naf}[1]{\ensuremath{\mathrm{\nafsymbol}~#1}}

\newcommand{\head}{\func{head}}
\newcommand{\body}{\func{body}}
\newcommand{\facts}{\func{facts}}
\newcommand{\bodyp}{\func{body^+}}
\newcommand{\bodyn}{\func{body^-}}
\newcommand{\ground}{\ensuremath{\func{grd}}}

\newcommand{\vars}{\ensuremath{\func{vars}}}

\usepackage{csquotes}
\usepackage{pdfpages}
\usepackage{graphicx}
\usepackage{tikz}

\usepackage{mathtools}
\usepackage{algpseudocodex}
\usepackage{float}

\usepackage{multicol}
\usepackage{booktabs} 

\ifpdf
  \usepackage{underscore}         
  \usepackage[T1]{fontenc}        
\else
  \usepackage{breakurl}           
\fi

\title{Investigating the Grounding Bottleneck \\ 
for a Large-Scale Configuration Problem: \\Existing Tools and Constraint-Aware Guessing}
\author{Veronika Semmelrock
\institute{Department of Artificial Intelligence and Cybersecurity\\ University of Klagenfurt, Austria}
\email{veronika.semmelrock@aau.at}
\and
Gerhard Friedrich\thanks{This research was funded by the Austrian Science Fund (FWF) 10.55776/COE12 and by the FFG Austria (\url{www.ffg.at}) project SAELING.}
\institute{Department of Artificial Intelligence and Cybersecurity\\ University of Klagenfurt, Austria}
\email{gerhard.friedrich@aau.at}
}
\def\titlerunning{Investigating the Grounding Bottleneck}
\def\authorrunning{V. Semmelrock \& G. Friedrich}
\begin{document}
\maketitle

\newtheorem{lemma}{Lemma}[section]
\newtheorem{proposition}{Proposition}

\begin{abstract}
Answer set programming (ASP) aims to realize the AI vision: The user specifies the problem, and the computer solves it. Indeed, ASP has made this vision true in many application domains. However, will current ASP solving techniques scale up for large configuration problems? As a benchmark for such problems, we investigated the configuration of electronic systems, which may comprise more than 30,000 components. We show the potential and limits of current ASP technology, focusing on methods that address the so-called grounding bottleneck, i.e., the sharp increase of memory demands in the size of the problem instances. To push the limits, we investigated the incremental solving approach, which proved effective in practice. However, even in the incremental approach, memory demands impose significant limits. Based on an analysis of grounding, we developed the method \emph{constraint-aware guessing}, which significantly reduced the memory need.  
\end{abstract}

\section{Introduction}

Large-scale configurations of electronic systems may comprise more than 200 racks, 1,000 frames, 30,000 modules, 10,000 cables, and 2,000 other units \cite{Fle98}. Electronic systems are typically configured by placing modules into frames and frames into racks. Technical constraints specify which configurations are valid for given user requirements. These requirements may be defined by a set of key modules to be included in a configuration. A configuration specifies the needed components (e.g., racks, frames and modules) and their connections. 

Answer set programming (ASP) offers an expressive logical description language that allows a compact problem encoding using a fragment of first-order logic based on all-quantifiers. For a comprehensive definition of ASP's syntax and semantics, we refer to \cite{DBLP:journals/tplp/CalimeriFGIKKLM20}. Efficient solvers make ASP an almost perfect choice for implementing automated configuration systems that enable the user to specify the problem, and the computer solves it \cite{Fal18}. Problem instances of a problem are encoded as an answer-set program, and solutions to a problem instance are extracted from the answer sets ($\AS$s). 

Most ASP frameworks follow the so-called ground-and-solve approach. The first-order answer-set program is transformed to a quantifier-free representation (grounding step), and solving operates on this grounded version of the problem description. Since grounding may require super-linear space in the input size (typically defined by a set of facts), applying ASP can become unfeasible due to high memory demands - a phenomenon known as the grounding bottleneck.

Consequently, various methods for mitigating the grounding bottleneck exist, such as lazy grounding with domain-specific heuristics \cite{Com23}, compilation approaches \cite{Dod24}, or body-decoupled grounding \cite{Bei24}.

Given these proposals, an important question is if ASP is ready for large-scale configuration. In this paper, we report on our evaluation, showing the current status and limitations of the proposed approaches, even on elementary configuration problems of electronic systems. 

Interestingly, simple and effective heuristics are available for configuring such systems \cite{Fle96}. These heuristics are exploited for the incremental extension of partial solutions to a complete one, which was the key to applying AI technology for large-scale configuration. Consequently, we investigated the potential of ASP solvers in the convenient case where current ASP solvers can employ such heuristics to extend partial solutions incrementally. 

Based on an analysis of the grounded problem instances, we observe that current methods allow choices that are obviously inconsistent and result in significant memory demands. Therefore, we propose the technique of \emph{constraint-aware guessing}, which saves memory by orders of magnitude. Both \emph{constraint-aware guessing} and the compilation approach of \cite{Dod24} substantially increase the efficiency of the incremental generation of solutions. 


The paper is structured as follows. We describe a configuration problem comprising the basic elements of electronic systems in Sec. \ref{sec:HCP}. Based on this problem, we analyze current ASP tools and their potential to solve large-scale configurations in Sec. \ref{sec:benchmarksStateOfTheArt}. In Sec. \ref{sec:incrementalSolving} we then outline an incremental approach to investigate the limits under the favorable assumption that partial solutions can be incrementally extended. In Sec. \ref{sec:CAG} we introduce the method \emph{constraint-aware guessing} and show its application to the introduced configuration problem. The results of evaluating the incremental approach are presented in Sec. \ref{sec:benchmarksIncrementalSolving} followed by our conclusions.

\section{Configuration of electronic systems} \label{sec:HCP}
To introduce the problem of configuring electronic systems, Siemens formulated the technology-\newline independent \emph{house configuration problem (HCP)} \cite{Rya11}.
In the HCP, the configuration of a house, which includes the entities things (i.e., modules), persons, cabinets (i.e., frames), and rooms (i.e., racks), is required. Existing things and persons, as well as the ownership relation of things belonging to a person is provided as input via \texttt{person/1}, \texttt{thing/1} and \texttt{personTOthing/2} facts, where each person can own any number of things, but each thing belongs to only one person. Domains for the possible rooms and cabinets that can exist are provided as input via \texttt{roomDomain/1} and \texttt{cabinetDomain/1} facts. The output, i.e., a configuration, is represented by a containment of things in cabinets and cabinets in rooms, as expressed by \texttt{cabinetTOthing/2} and \texttt{roomTOcabinet/2}, and constrained by specific requirements.  

The requirements for a valid configuration are the following: (1) each thing must fit into a cabinet; (2) the maximum capacity of any cabinet is five things; (3) each cabinet must be placed in a room; (4) the maximum capacity of any room is four cabinets; (5) each cabinet belongs to only one person and (6) each room belongs to only one person. Additionally to the standard description of HCP, things and cabinets are numbered, and a thing can only be assigned to a specific cabinet if no higher-numbered cabinet holds a lower-numbered thing.
The respective ASP coding is as follows ($``\texttt{:-}"$ is $``\leftarrow"$):

\begin{footnotesize}
\begin{verbatim}
cabinet(C)   :- cabinetDomain(C), not cabinet_n(C).
cabinet_n(C) :- cabinetDomain(C), not cabinet(C).

room(R)   :- roomDomain(R), not room_n(R).
room_n(R) :- roomDomain(R), not room(R).

cabinetTOthing(C,T)   :- thing(T), cabinetDomain(C), not cabinetTOthing_n(C,T).
cabinetTOthing_n(C,T) :- thing(T), cabinetDomain(C), not cabinetTOthing(C,T).
:- 1 > #count { C : cabinetTOthing(C,T) }, thing(T). %% Req. 1
:- cabinetTOthing(C1,T), cabinetTOthing(C2,T), C1 < C2. %% Req. 1
:- 6 <= #count { T : cabinetTOthing(C,T), thing(T) }, cabinet(C). %% Req. 2
:- cabinetTOthing(C1,T1), cabinetTOthing(C2,T2), C1 < C2, T1 > T2. %% Additional ordering req.


roomTOcabinet(R,C) :- cabinet(C), roomDomain(R), not roomTOcabinet_n(R,C).
roomTOcabinet_n(R,C) :- cabinet(C), roomDomain(R), not roomTOcabinet(R,C).
:- 1 > #count { R : roomTOcabinet(R,C) }, cabinet(C). %% Req. 3
:- roomTOcabinet(R1,C), roomTOcabinet(R2,C), R1 < R2. %% Req. 3
:- 5 <= #count { C : roomTOcabinet(R,C), cabinetDomain(C) }, room(R). %% Req. 4

personTOcabinet(P,C) :- personTOthing(P,T), cabinetTOthing(C,T). %% Req. 5
:- personTOcabinet(P1, C), personTOcabinet(P2, C), P1 < P2. %% Req. 5
personTOroom(P,R) :- personTOcabinet(P,C), roomTOcabinet(R,C). %% Req. 6
:- personTOroom(P1,R), personTOroom(P2,R), P1 < P2. %% Req. 6

room(R1) :- roomDomain(R1), roomDomain(R2), room(R2), R1 < R2.
cabinet(C1) :- cabinetDomain(C1), cabinetDomain(C2), cabinet(C2), C1 < C2.

room(R) :- roomTOcabinet(R,C).
cabinet(C) :- cabinetTOthing(C,T).
\end{verbatim}
\end{footnotesize}
As an example, the stated configuration can be created for the following input data:
\begin{footnotesize}
\begin{verbatim}
%% input facts
person(1). thing(1). thing(2). personTOthing(1,1). personTOthing(1,2).
person(2). thing(3). thing(4). personTOthing(2,3). personTOthing(2,4).
roomDomain(1). roomDomain(2). cabinetDomain(1). cabinetDomain(2).
%% output facts, representing a configuration  
cabinetTOthing(1,1). cabinetTOthing(1,2). roomTOcabinet(1,1).
cabinetTOthing(2,3). cabinetTOthing(2,4). roomTOcabinet(2,2).
\end{verbatim}
\end{footnotesize}

The presented encoding deliberately avoids the use of choice rules and disjunctive rule heads. This syntactic restriction makes the encoding compatible with various systems that do not support such rules.

\section{Benchmarking grounding strategies in ASP: An evaluation of existing methods}
\label{sec:benchmarksStateOfTheArt}

We investigated the latest methods for mitigating the grounding bottleneck, such as lazy grounding combined with domain-specific heuristics \cite{Com23}, compilation approaches \cite{Dod24}, or body-decoupled grounding \cite{Bei24}. To get the full picture of current approaches, we included the tools \emph{clingo} \cite{potasscoguide} and \emph{DLV} \cite{Adr18} as representatives for the standard ground-and-solve approach.

\vspace{4pt} \noindent\emph{Gringo and I-DLV (ground-and-solve):}
 In clingo, the application of the grounder \emph{gringo} is followed by the usage of the solver \emph{clasp}, while the system DLV uses the solver \emph{WASP} after applying \emph{I-DLV} for grounding. Due to I-DLV performing well in some benchmarks, we also evaluated the combination of I-DLV \cite{Cal17} as the groudner with clasp as the solver.

\vspace{4pt} \noindent\emph{Alpha (lazy grounding with domain-specific heuristics):}
\emph{Alpha} \cite{Com23} is a \emph{lazy-grounding system} that circumvents the grounding bottleneck by interleaving the grounding and solving steps. In this interleaving phase, the ground instances of only those rules are instantiated, which are necessary by the solver's current state. 
Additionally, Alpha allows declaratively specified domain-specific heuristics to be included in the encoding to guide the solver and find $\AS$s faster. For the HCP simple and effective heuristics are available and the employed heuristics for our benchmarks can be found at \url{https://github.com/VeronikaSemmelrock/GroundingBottleneck}. 

\vspace{4pt} \noindent\emph{ProASP (compilation approach):}
\emph{ProASP} \cite{Dod24} is a system that partially overcomes the grounding bottleneck by skipping the grounding phase and instead adapting the solving phase via compiling in external propagators subprograms acting as constraints. Custom propagators generated based on the input encoding are injected in the CDCL solver GLUCOSE, which simulates the rules that are not grounded, such as variable initialization.  
ProASP implements mechanisms to merge the grounding and compilation approaches by allowing the user to specify which parts of the encoding should be compiled or grounded. \cite{Dod24} reports on six different ways to split an encoding by rule type. After testing each possible split with the HCP encoding, using medium-sized instances of 200 things, the constraints split gave the optimal performance w.r.t. runtime and memory consumption. Consequently, in the ProASP experiments, all constraints that do not contain aggregates are compiled, while the remaining rules are grounded.


\vspace{4pt} \noindent\emph{Newground (body-decoupled grounding):}
\emph{Newground} \cite{Bei24} tackles the grounding bottleneck through program rewriting and hybrid-grounding by applying body-decoupled grounding (BDG) to parts of a program. Newground is a system that improves the performance of a traditional ground-and-solve system by applying BDG to rules with many variables, shifting the effort from the grounder to the solver. In BDG, body elements are grounded independently. Because the approach is outperformed by traditional grounding on small rule bodies, newground implements mechanisms for the user to apply BDG on dense, tight rules only while the rest of the encoding is ground, which is called hybrid-grounding. In our experiments, newground is applied exclusively to the constraint \texttt{:- cabinetTOthing(C1,T1), cabinetTOthing(C2,T2), C1 < C2, T1 > T2.}, which turned out to be the best hybrid-grounding split. Note that all other rules in the HCP contain fewer variables in the body.

\vspace{4pt} \noindent\emph{Local search} is one of the most successful search strategies if memory consumption of search algorithms is an issue. In \cite{Eit24}, the authors describe methods to apply ASP for improving solutions using local search. They assume that an initial solution is provided, e.g., via greedy search. Note that we aim to realize the vision that the user only specifies the problem; thus, this specification should be sufficient to generate an initial solution. 
Since we seek to find a first solution for the HCP, we refrain from an evaluation of \cite{Eit24}. 

\vspace{4pt} \noindent\emph{Experimental setup:}
To evaluate the performance of state-of-the-art ASP tools under increasing grounding demands, a structured benchmarking setup was developed, focusing on systematically scaling HCP instances. The HCP domain serves as a simplified lower bound for the complexity of typical large-scale configuration problems. Instances were generated synthetically by incrementally adding five persons per step, each with 10 associated things, while proportionally expanding the cabinet and room domains (two cabinets and one room per person) to maintain consistency. This construction was implemented in favor of ASP by adding the exact amount of required cabinets and rooms. One test consisted of searching for the first $\AS$ of a problem instance of the HCP with a timeout of one hour. This setup allows for controlled growth in problem size, as each person contributes a constant and known increase in the number of things, while keeping the encoding logic unchanged, enabling analysis of execution time and memory consumption under realistic scaling. Note that things correspond to key modules as described in \cite{Fle98}. These key modules represent the user requirements which are the input for generating configurations.

All experiments were executed on a VM with exclusive access to 32 GB of RAM and four cores of an AMD EPYC 7H12 @ 2.6GHz CPU. The VM was running Ubuntu 22.04.4 LTS x86\_64 and had 986 GB of disk space.  Tested systems and combinations are Alpha, clingo (gringo and clasp), DLV (I-DLV and WASP), I-DLV+clasp, ProASP and newground. Versions of systems are clingo 5.7.1, DLV 2.1.2, newground version NaGG (newground2, IJCAI24)\footnote{Available at: \url{https://github.com/alexl4123/newground/tree/ijcai24-NaGG}} and Alpha from \cite{Com23}\footnote{Available at: \url{https://git-ainf.aau.at/DynaCon/website/tree/master/supplementary_material/2022_JAIR_Domain-Specific_Heuristics}}. The version of ProASP was retrieved from GitHub\footnote{Available at: \url{https://github.com/MazzottaG/ProASP}} on 16-12-2024. This is the latest available release at the time of writing. 

\vspace{4pt} \noindent\emph{Analysis:}
\begin{figure}[ht]
    \centering
    \includegraphics[width=\textwidth]{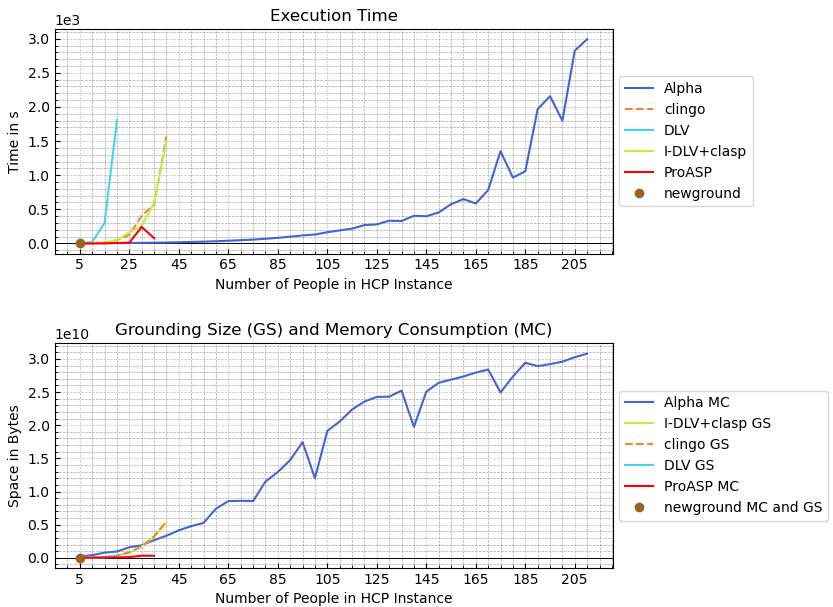}
    \caption{Performance of state-of-the-art systems}
    \label{fig:stateOfTheArtTools}
\end{figure}
The experimental evaluation (reported in Fig. \ref{fig:stateOfTheArtTools}) shows that within the time limit of one hour at most configurations for approx. 2,100 things can be generated by lazy grounding, whereas the standard ground-and-solve approach reaches a maximum of 400 things. DLV, clingo, and I-DLV+clasp show a steep increase in the grounding size. The large size of the grounding led to an error in clasp. 

Utilizing its compilation approach, ProASP demonstrates a slightly lower capacity, achieving an instance size of 350 within the same timeframe. However, as the second plot illustrates, the grounding size of ProASP remains comparatively low in relation to the other approaches. Yet, ProASP encounters challenges regarding time efficiency, which hinders its ability to solve larger instances. Based on the information provided by the current version of ProASP, it is currently not possible for the user to determine the reasons for exceeding the time limit.     

Newground is only capable of solving the first instance of 50 things and the next instance size of 100 things times out during solving. This limitation can be attributed to the substantial time consumption of the solving step, as the application of newground moves effort from the grounder to the solver.

Note that real-world problem instances may comprise 30,000 things and a much more sophisticated problem description. Although our evaluation favors applying current ASP technology for configuring electronic systems following the module/frame/rack structure, the configuration of large real-world problem instances needs improvements by orders of magnitude. 

\section{Incremental solving} 
\label{sec:incrementalSolving}

Our investigation shows that using current technology, a one-step solving approach that passes the complete problem instance to an ASP solver does not scale to the required problem instance size, not even in the simplified HCP encoding. 
Siemens applies an incremental extension of partial solutions to a full solution  \cite{FalknerFHSS16}. To explore the boundaries, we will again assume a favorable case for ASP tools.

In this favorable scenario, we suppose that a partial solution can be extended to a complete solution and that we make a perfect guess which of the input facts must be added such that no backtracking is required, i.e., no parts of a partial solution must be retracted for generating a complete solution. We assume that a problem is defined by an ASP program $P$ and a set of indexed sets of input facts $I = \{ I_1, \ldots, I_n \}$. A solution to the problem is an $\AS$ $\Delta$ of $P \cup \bigcup_{i} I_i$.

To generate a solution, we use function \textsc{IncrementalSolving} (see below). Function $\facts(\interpretation)$ transforms an interpretation (rsp. $\AS$) $\Delta$ to a set of facts by converting each atom of $\Delta$ to a fact. 
\begin{figure}[h]
\begin{algorithmic}
\Function{IncrementalSolving}{$P$,$\{ I_1, \ldots, I_n \}$}
\LComment{Inputs: an ASP program $P$, indexed sets of input facts $\{ I_1, \ldots, I_n \}$ }
\LComment{Output: an answer set $\Delta$ of $P \cup \bigcup_{i} I_i$ }
\State $\Delta \gets \emptyset$
\State $\textit{InputFacts} \gets \emptyset$
\For{$i = 1, \dots, n$}
\State $\textit{InputFacts} \gets \facts(\Delta) \cup I_i$
\State $\Delta \gets \textit{ASPsolver}(P \cup \textit{InputFacts})$ \Comment{An ASP solver outputs an $\AS$; input is an ASP program}
\EndFor
\Return $\Delta$
\EndFunction
\end{algorithmic}
\label{fig:incremental}
\end{figure}

However, the grounding of an (incrementally extended) ASP program may result in guessing, in which inconsistencies are easy to recognize. Avoiding inconsistent guesses can significantly reduce the size of the grounded program. We illustrate this with a simple example, where modules must be placed in exactly one frame, and frames have a capacity of one module. 

\begin{footnotesize}
\begin{multicols}{2}
\begin{verbatim}
%% input facts
frame(1). frame(2). 
module(1). module(2). 
mINf(1,2).
%% modules are placed into frames 
%% rule r1
mINf(X,Y) :- not mINf_n(X,Y), 
    module(X), frame(Y).
%% rule r2
mINf_n(X,Y) :- not mINf(X,Y),
    module(X), frame(Y).

%% c1: a frame can hold 1 module
:- module(M1), module(M2), frame(F), 
    mINf(M1,F), mINf(M2,F), M1 <> M2.
%% c2: a module can only be placed in 1 frame
:- module(M), frame(F1), frame(F2), 
    mINf(M,F1), mINf(M,F2), F1 <> F2.

%% modules must be placed in frames
modulePlaced(X) :- module(X), mINf(X,Y).
:- module(X), not modulePlaced(X).
\end{verbatim}
\end{multicols}
\end{footnotesize}
The grounder \emph{gringo} \cite{potasscoguide} produces the following ground constraints and ground rules for guessing (we omit input facts and duplicate rules for presentation purposes):
\begin{footnotesize}
\begin{multicols}{2}
\begin{verbatim}
%%% guessing
mINf(1,1) :- not mINf_n(1,1).
mINf(2,1) :- not mINf_n(2,1).
mINf(2,2) :- not mINf_n(2,2).
mINf_n(1,1 ):- not mINf(1,1).
mINf_n(2,1) :- not mINf(2,1).
mINf_n(2,2) :- not mINf(2,2).
%%% checking
modulePlaced(1).
modulePlaced(2):-mINf(2,1).
modulePlaced(2):-mINf(2,2).
:- not modulePlaced(2).
:- mINf(1,1).
:- mINf(2,1),mINf(2,2).
:- mINf(2,2),mINf(2,1).
:- mINf(1,1),mINf(2,1).
:- mINf(2,1),mINf(1,1).
:- mINf(2,2).
\end{verbatim}
\end{multicols}
\end{footnotesize}

Note, that \texttt{module(1)} is already placed in \texttt{frame(2)}. Asserting \texttt{mINf(1,1)} or \texttt{mINf(2,2)} by guesses will lead to violated constraints. If we avoid these guesses that will not lead to a consistent solution, we can remove grounded constraints and rules.   

\section{Constraint-Aware Guessing (CAG)}
\label{sec:CAG}

In this section, we introduce the method for \emph{constraint-aware guessing} to filter guesses to reduce the size of grounded programs. Note that problem instances are encoded by a program $\prog$ and every $\AS$ corresponds to a solution. We assume a partitioning of $\prog = \prog^{check} \cup \prog^{guess}$ s.t. for every interpretation $\interpretation$ of $\prog$ which does not correspond to a solution, $\prog^{check} \cup \facts(\interpretation)$ is unsatisfiable.

Step 1: Let rule $r$ be a rule of $\prog^{guess}$ with a nonempty head whose body contains default negated atoms, i.e., $\bodyn(r) \neq \emptyset$. Consequently, this rule is subject to a guess. Let $c$ be a constraint of $\prog$ where the atom $a_r$ in $\head(r)$ unifies with an atom $a_c$ in $\bodyp(c)$ s.t. $a_r = a_c \gamma$ where $\gamma$ is a substitution that maps terms to terms (e.g., variables to variables). We assume w.l.o.g. that if atoms $a_r, a_c$ are atoms with arguments, these arguments are variables. Arguments that are not variables (e.g., constants) can be emulated by adding equality atoms to the body. In addition, we assume ground predicates bind the variables in $\bodyp(c)$ to assure safety, and the names of variables in $r$ and $c$ are different.

For each atom $a_c \in  \bodyp(c)$ where  $a_r = a_c \gamma$ we generate a \emph{filter condition} $b(r,a_c)$ $ =$ $ ($\body(c)$ - $ $a_c) \gamma$. For constraint \texttt{c1} and rule \texttt{r1} in our example this results in

\begin{tabbing}
    $b_1:$~ \=  $\{$ \texttt{module(X), module(M2), frame(Y), mINf(M2,Y), X <> M2} $\}$ \kill
    $b_1:$~ \>  $\{$ \texttt{module(X), module(M2), frame(Y), mINf(M2,Y), X <> M2} $\}$ \\
    $b'_1:$~ \>  $\{$ \texttt{module(M1), module(X), frame(Y), mINf(M1,Y), M1 <> X} $\}$
\end{tabbing}
Note $b_1$ and $b'_1$ are equivalent, aside from renaming variables \texttt{M1, M2}. Therefore, we consider only $b_1$ subsequently.
For constraint \texttt{c2} in our example we generate
\begin{tabbing}
    $b_2:$~ \=  $\{$ \texttt{module(X), frame(Y), frame(F2), mINf(X,F2), Y <> F2} $\}$ \kill
    $b_2:$~ \>  $\{$ \texttt{module(X), frame(Y), frame(F2), mINf(X,F2), Y <> F2} $\}$
\end{tabbing}
As for constraint \texttt{c1}, we do not consider the equivalent filter condition. 

Step 2: Observe that if condition $b_1 \gamma$ and $\body(\texttt{r1})  \gamma$ are true in an interpretation $\interpretation$ for a substitution $\gamma$ then constraint \texttt{c1} is unsatisfiable. 

\vspace{4pt} Provided $\prog$ contains a program $\prog^{check}$,  we can add the negation of all filter conditions to the body of $r$, which does not reduce the set of $\AS$s of $\prog$ but reduces the ground instantiations of $r$. Let $B(r,\prog)$ be all the filter conditions for a rule $r$ for all the constraints $c \in \prog$. We replace $r$ by its filtered version $r^f$ defined as $\head(r) \texttt{:-} \body(r) \bigcup_{b(r,a_c) \in B(r, \prog)}  \{\naf b(r,a_c)\}.$ Since the default negation is not applicable for a set of atoms, we employ the expression 
$\texttt{ \#count \{1:} b(r,a_c) \texttt{\}} $ $< 1$ for  $\naf b(r,a_c)$ if the filter condition $b(r,a_c)$  does not contain aggregates. Note that syntactically it is not possible for aggregates to contain aggregates. If $b(r,a_c)$ is a single aggregate atom, we can negate the aggregate function, e.g., replacing $<$ by $\geq$. If $b(r,a_c)$ comprises multiple atoms where at least one atom is an aggregate atom, we could introduce auxiliary rules and atoms to move an aggregate outside an aggregate. However, this results in an increase of the size of the grounded program.

\vspace{4pt} \noindent\emph{Proposition:}
Let $r$ be a rule with $\bodyn(r) \neq \emptyset$, $r^f$ its filtered version, $\prog^{check} = \prog \setminus \prog^{guess}$ a program as described, and  program $\prog^f = ( \prog \setminus \{ r \}) \cup \{ r^f \}$ (i.e., rule $r$ is exchanged by $r^f$). Every $\AS$ (solution) of $\prog$ is an $\AS$ (solution) of $\prog^f$. By the properties of $\prog^{check}$ all $\AS$s of $\prog^f$ correspond to a solution.

\emph{Proof:} For every $\AS$ of $\prog$ where $\body(r) \sigma$ is true for some $\sigma$, $(b(r,a_c)\sigma)\delta$ must be false for all possible ground substitutions $\delta$. Otherwise a constraint would be unsatisfied. Consequently, we can extend $r$ to $r^f: \head(r) \texttt{:-} \body(r) \cup \{ not \ b(r,a_c)\}.$ and $\body(r^f)$ is true in $\AS$. Therefore, if $\interpretation$ is an $\AS$ for $P$ then $\interpretation$ is an $\AS$ for $\prog^f$. However, $\prog^f$ may comprise $\AS$s which are not $\AS$s of $\prog$.  Yet, $\prog^{check}$ assures that these $\AS$s correspond to solutions. $\Box$

In our example, we extend rule \texttt{r1} with constraint-aware filter conditions as follows:
\begin{verbatim}
mINf(X,Y) :- not mINf_n(X,Y), module(X), frame(Y),
	#count{1: module(X), module(M2), frame(Y), mINf(M2,Y), X <> M2 } < 1,
	#count{1: module(X), frame(Y), frame(F2), mINf(X,F2), Y <> F2 } < 1.
\end{verbatim}
This results in a significantly smaller grounding generated by \emph{gringo} for the guessing and checking part of the grounded program (input facts are omitted):
\begin{footnotesize}
\begin{multicols}{2}
\begin{verbatim}
%%% guessing
mINf(2,1):- not mINf_n(2,1),
    #delayed(1),#delayed(2).
mINf_n(1,1).
mINf_n(2,1):- not mINf(2,1).
mINf_n(2,2).
#delayed(1) <=> #count{}
#delayed(2) <=> #count{}
%%% checking
:- not modulePlaced(2).
modulePlaced(1).
modulePlaced(2):- mINf(2,1).
\end{verbatim}
\end{multicols}
\end{footnotesize}

The delay atoms are added by \emph{gringo} as auxiliary constructs for dealing with aggregates. 

Note that it is possible to generate new constraints by replacing atoms $a_c$ in a constraint with bodies of rules whose head $a_r$ matches $a_c$, allowing additional filter conditions to be generated. For the HCP we consider the follwoing generated constraint
\begin{footnotesize}
\begin{verbatim}
:- personTOthing(P1,T1), cabinetTOthing(C,T1),  
   personTOthing(P2,T2), cabinetTOthing(C,T2), P1 < P2.
\end{verbatim}
\end{footnotesize}

 \noindent\emph{Applying constraint-aware guessing to HCP:}
To demonstrate and evaluate the proposed method, which constitutes a general strategy for rewriting ASP encodings, it was applied to the HCP as a representative case study. The complete resulting encoding of adding filter conditions to HCP rules can be found in \url{https://github.com/VeronikaSemmelrock/GroundingBottleneck}. As an example, the rule for guessing \texttt{cabinetTOthing/2} is extended as follows: 

\begin{footnotesize}
\begin{verbatim}
cabinetTOthing(C,T) :- thing(T), cabinetDomain(C), not cabinetTOthing_n(C,T),  
    #count{1: cabinetTOthing(C1,T), C < C1 } <1, %% Req. 1
    #count{1: cabinetTOthing(C1,T),  C1 < C } <1, %% Req. 1
    #count{1: cabinetTOthing(C1,T1), C1 < C, T1 > T } <1, %% Additional ordering requirement
    #count{1: cabinetTOthing(C1,T1), C < C1, T > T1 } <1, %% Additional ordering requirement
    #count{1: personTOthing(P1,T), personTOthing(P2,T2), cabinetTOthing(C,T2), P1<P2}<1, %Req. 5
    #count{1: personTOthing(P1,T1), personTOthing(P2,T), cabinetTOthing(C,T1), P1<P2}<1. %Req. 5
\end{verbatim}
\end{footnotesize}

This rule uses multiple aggregates derived from constraints of mentioned requirements, enforcing consistency in assignments and ordering between things, cabinets, and persons. Note that both derived \texttt{\#count}-aggregates of requirement 1 combined behave like \texttt{\#count\{1: cabinetTOthing(C1,T), C != C1\} < 1}, which ensures that each thing $T$ is placed into at most one cabinet. The two \texttt{\#count}-aggregates of requirement 5 are created via substitution and combination of the two rules of requirement 5 of the unchanged HCP encoding (see Section \ref{sec:HCP}).

\section{Benchmarking incremental solving} 
\label{sec:benchmarksIncrementalSolving}

The implementation of the function \textsc{IncrementalSolving} (see Section \ref{sec:incrementalSolving}) employs an ASP solver. Each invocation of the ASP solver is stateless; that is, no information is retained between consecutive calls and the solver is reinitialized from scratch for every run. In our evaluations, we applied clingo, DLV, newground, and ProASP as well as the combination of I-DLV and clasp, resulting in \emph{incremental-clingo}, \emph{incremental-DLV}, \emph{incremental-newground}, \emph{incremental-ProASP}, and \emph{incremental-I-DLV+clasp}.
The implementation of CAG for HCP builds on the incremental solving system incremental-clingo. In contrast to incremental-clingo, which operates on the HCP encoding of Section \ref{sec:HCP}, CAG employs the adapted encoding including filter conditions, detailed in Section \ref{sec:CAG}. 

An attempt was made to combine the CAG approach with incremental-ProASP; however, the integration failed due to ProASP's syntactic limitations, particularly its handling of aggregates. These issues prevented correct parsing and compilation, making the combination infeasible in its current form. 

\vspace{4pt}\noindent\emph{Experimental setup:}
The experimental setup for incremental solving differs from that in Section \ref{sec:benchmarksStateOfTheArt} solely regarding the tested systems. 
The function \textsc{IncrementalSolving} was implemented via a Python script. Initially, the script generates all the facts for a problem instance (as in Section \ref{sec:benchmarksStateOfTheArt}), but subsequently divides them into fact batches. These fact batches then serve as the input facts $I = \{ I_1, \ldots, I_n \}$, which are added in each iteration of the incremental solving process to extend the solution. In the HCP, it is possible to configure all the things of one person independently from all other people's facts; thus, a fact batch can contain facts of one person up to several people, i.e., the \texttt{person/1}, \texttt{thing/1}, and \texttt{personTOthing/2} facts as well as the extensions of the domains via \texttt{roomDomain/1} and \texttt{cabinetDomain/1} facts. 
The Python script implements code to call the solvers, providing the fact batch and the problem encoding, and extracting the found $\AS$. Next, this $\AS$ is parsed and added as facts to the fact batch of the next iteration. 
The decomposition of the problem instance into fact batches introduces a variable \texttt{persons-per-iteration} \texttt{(PPI)}, which describes the number of people whose facts are added to one fact batch for one iteration.
All tools were initially tested using a \texttt{PPI} of five, as this showed the best performance in CAG. The optimal \texttt{PPI} was determined for instances up to 100 people. In this setting, a \texttt{PPI} of five is also optimal for incremental-ProASP. Employing ProASP for incremental solving, the constraints split was again identified as the most efficient and least memory-consuming approach. 
Furthermore, incremental-clingo, which is after CAG the most effective ground-and-solve system, was additionally optimized by determining an optimal \texttt{PPI}. The optimized \texttt{PPI}  for incremental-clingo was determined to be one which then was employed in a separate benchmark test, following the same principles. We denote these results in our graphs by \texttt{PPI-opt. incremental-clingo}.

\begin{figure}
    \centering
    \includegraphics[width=\textwidth]{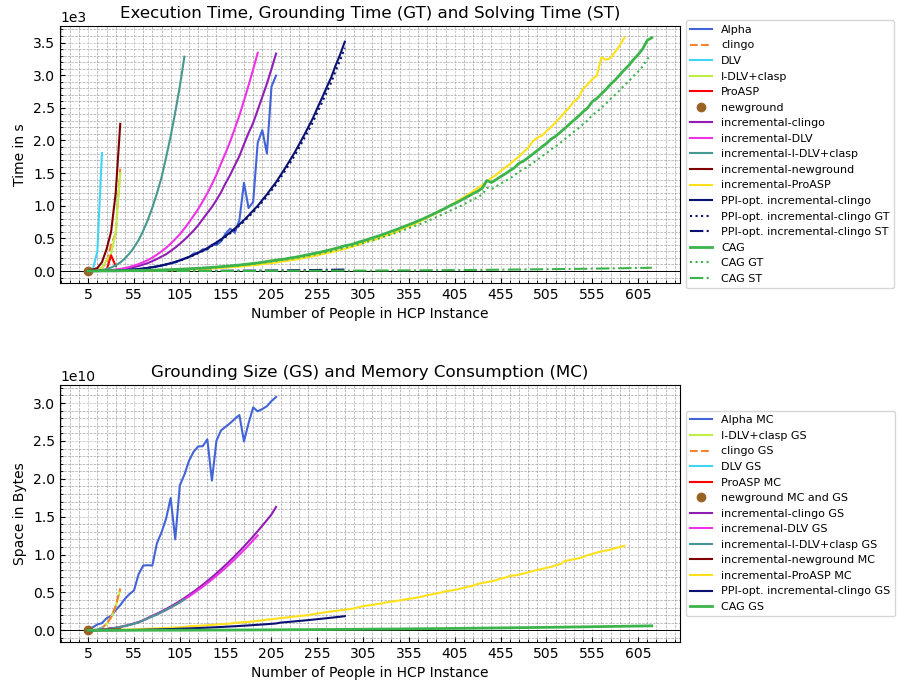}
    \caption{Benchmarks of all tools and approaches}
    \label{fig:allApproaches}
\end{figure}

\vspace{4pt} \noindent\emph{Comparative analysis:}
The plots in Figure \ref{fig:allApproaches} demonstrate the results of the state-of-the-art tools and systems and all incremental solving tools, including CAG. The results show that CAG solves the largest instances, followed by incremental-ProASP.

There was a substantial improvement in the incremental application of clingo, DLV, I-DLV+clasp, newground, and ProASP compared to their one-step application.
The most significant relative enhancement was observed in ProASP, where the incremental application of ProASP on the HCP successfully solved a 1.585 \% larger instance within one hour. Due to the limited transparency of the ProASP's internal mechanisms, it is challenging to precisely attribute the source of performance improvements observed when it is applied incrementally. The second largest relative improvement was observed with DLV. Newground also exhibited a substantial improvement by 700~\% being able to solve a maximum of 400 things. In comparison, the incremental application of clingo on the HCP resulted in a 425 \% improvement in instance sizes solved, while incremental-I-DLV+clasp reached 175 \% larger instances. 

Furthermore, incremental-clingo is depicted in the plot once with a PPI of 5, like all other approaches, and once with its optimized PPI value of 1. The optimization of the PPI value for incremental-clingo resulted in the approach being able to solve an instance with 2,850 things instead of 2,100. Note that the second plot in Figure \ref{fig:allApproaches} depicts the grounding size in the final iteration, which is the largest during the entire execution. In the case of incremental-clingo, this illustrates the effort involved in adding the facts of five individuals, each introducing 10 things. Whereas the PPI-opt. incremental-clingo values show the effort of adding the facts of only one person. Thus, the lines of incremental-clingo are not comparable. 

A comparison between the two best performing ground-and-solve approaches, PPI-opt. incre\-mental-clingo and CAG, shows a performance advantage of 118 \% for CAG, where PPI-opt. incremental-clingo can solve an instance size of 2,850 things in one hour, while CAG can solve an instance size of 6,200 things. It is possible to compare the grounding size of incremental-clingo and CAG, as both have a PPI of 5. This shows a 99.6 \% reduction in the grounding size of CAG in the last iteration of the largest common instance of 2,100 things. A naive comparison of PPI-opt. incremental-clingo and CAG shows that the grounding size of CAG is 93.1 \% smaller when comparing the largest common instance although CAG grounds five times as many new facts in the last iteration as PPI-opt. incremental-clingo. 

A comparison of the two approaches that demonstrated the highest overall performance, CAG and incremental-ProASP, revealed that their performance was comparable despite the difference in their underlying mechanisms. Incremental-ProASP reached an instance size of 5,900, while CAG successfully solved 6,200 things within one hour. As both have an optimal PPI of 5, comparing their performance regarding grounding size for CAG and memory consumption for incremental-ProASP is possible. In their largest common instance of 5,900 things, CAG has a grounding size that is 95 \% smaller than the memory consumption of incremental-ProASP. This clearly demonstrates the advantage of CAG regarding memory demands when instance sizes are increased. 

Moreover, the first plot in Fig. \ref{fig:allApproaches} includes separate metrics of total grounding time (GT) and solving time (ST) for the CAG approach, which proved to be the best approach overall, and the PPI-opt. incremental-clingo approach, which was the second best ground-and-solve approach. CAG solves the largest instances and improves all performance metrics by lowering execution, grounding, and solving time. In the CAG approach, approximately 94 \% of the execution time is allocated to grounding, while solving accounts for 1.5 \% of the total execution time across all instance sizes. The remaining 4.5 \% of the execution time is attributed to the Python script implementing the incremental solving mechanism. A similar trend is observed in the case of PPI-optimized incremental-clingo, with grounding accounting for approximately 97 \% of the total execution time. The remaining 3 \% is divided into approx. 0.6 \% solving time and approx. 2 \% time for executing the Python script.

\vspace{4pt} \noindent\emph{Other methods for incremental solving and problem decompositions:}
In \cite{Geb19} multi-shot ASP solving was introduced to deal with continuously changing programs. Multi-shot ASP solving aims to avoid redundancies in relaunching grounder programs and benefits from the solver’s learning capacities. In our incremental solving process, we extend program $P_k = P \cup \bigcup_{i = 1 \ldots k} I_i$ by input facts $I_{k+1}$ and an $\AS$  $\Delta_k$ for $P_k$. Consequently, all grounded rules of $P_k$ are deleted or reduced to facts, and a new problem is solved. Therefore, the benefits of reusing the grounding of $P_k$ by multi-shot ASP are limited. Adding $\Delta_k$ to $P_k$ by facts is impossible because this extension is not compositional, i.e., an atom would be defined in two different modules. Adding $\Delta_k$ as externals has no effect because the atoms in $\Delta_k$ are output atoms in $P_k$. It is possible to add $\Delta_k$ to $P_k$ via constraints, however, this entails a substantial reformulation of the problem encoding, which is preferably done automatically. This reformulation is beyond the scope of this investigation and is part of future work.

In \cite{Abs14} an approach (D-FLAT) based on a tree-decomposition and dynamic programming employing ASP was introduced. Dynamic programming exploits a tree structure of the problem. It assumes that (partial) solutions for subsets of nodes in a graph can be combined so that a global solution can finally be generated by combining the partial solutions. Consequently, for a given problem, the design of the appropriate data structures and answer-set programs employed in dynamic programming is required. The transformation from ASP to SAT (assuming appropriate syntactical restrictions of ASP) can realize the general applicability of dynamic programming for ASP because D-FLAT can solve SAT problems. However, such an approach does not apply to HCP for large instances because of the grounding size.

\section{Conclusion and discussion} 
We investigated the potential of current ASP methods for configuring large-scale electronic systems as described in \cite{Fle98}, which may comprise more than 30,000 modules, with a simplified HCP encoding serving as a lower bound for the complexities of industry problems, aiming to create a favourable case for ASP systems. For our evaluation, we defined a configuration problem that follows the standard module/frame/rack structure of such systems \cite{Rya11}. Still, we did not consider cables, different module types, or complex technical constraints (e.g., balancing physical quantities). Within a time limit of one hour, the most efficient ground-and-solve approach manages to configure systems in the size of approximately 400 modules (i.e., things in HCP). Due to the superlinear increase in space demands and runtime as the input size grows, polynomial regression predicts a grounding size of more than 5 exabytes and runtime of more than 10,000 years for 30,000 modules for the ground-and-solve approaches. A special case is lazy-grounding, which can push the limit to approx. 2,000 modules; however, effective heuristics must be available. We, therefore, investigated the case, which seems to be shared in many configuration problems \cite{Fle96,FalknerFHSS16}, where configuration problem instances can be incrementally solved. Incremental solving pushes the limit to approximately 3,000 modules for incremental-clingo and can be further improved by \emph{constraint-aware guessing}, which manages to double the size for the one hour time limit and has a significantly lower increase of memory demands (approx. 99\% savings). The incremental application of ProASP turned out to be comparable to CAG regarding the size of the solvable instances within a one hour time limit. However, memory demands of CAG are significantly lower compared to incremental-ProASP (approx. 95\% savings). 


Note that we assumed a configuration problem with few technical constraints and a low variety of component types. The challenge of solving such problems is the size of the problem instances and their solutions. Performing guesses that allow a solution is easy in many practical configuration problems \cite{FalknerFHSS16}. Whether current solving approaches will scale up for such domains or radical new approaches are required remains an essential open question.  

Future work includes a systematic investigation of grounding behavior in multi-shot ASP systems, similar to the analysis conducted in this work. Additionally, exploring how CSP solvers handle grounding for problems such as the HCP, which represents a lower bound in terms of complexity, may provide further insights into solver behavior and efficiency across paradigms.


\nocite{*}
\bibliographystyle{eptcs}
\bibliography{generic}
\end{document}